\pdfoutput=1

\documentclass[11pt]{article}

\usepackage[]{acl}

\usepackage{times}
\usepackage{latexsym}

\usepackage[T1]{fontenc}

\usepackage[utf8]{inputenc}

\usepackage{microtype}

\usepackage{xcolor}
\usepackage{multicol}
\usepackage{multirow}
\usepackage{booktabs}
\usepackage{graphicx}
\usepackage{amsmath}
\usepackage{amssymb}
\usepackage{float}
\usepackage{enumitem}
\newcommand{\abs}[1]{\left \lvert #1 \right \rvert}
\newcommand\gn[1]{
}

\title{Breaking Down Multilingual Machine Translation}

\author{Ting-Rui Chiang$^1$ \quad Yi-Pei Chen$^2$ \quad Yi-Ting Yeh$^1$ \quad Graham Neubig$^1$ \\
    $^1$Carnegie Mellon University, $^2$The University of Tokyo \\
    \texttt{\{tingruic,yitingye,gneubig\}@cs.cmu.edu} \\
    \texttt{ypc@g.ecc.u-tokyo.ac.jp }
}

\date{}

\begin{document}
\maketitle

\begin{abstract}
While multilingual training is now an essential ingredient in machine translation (MT) systems, recent work has demonstrated that it has different effects in different multilingual settings, such as many-to-one, one-to-many, and many-to-many learning.
These training settings expose the encoder and the decoder in a machine translation model with different data distributions.
In this paper, we examine how different varieties of multilingual training contribute to learning these two components of the MT model. 
Specifically, we compare bilingual models with encoders and/or decoders initialized by multilingual training. 
We show that multilingual training is beneficial to encoders in general, while it only benefits decoders for low-resource languages (LRLs). 
We further find the important attention heads for each language pair and compare their correlations during inference.
Our analysis sheds light on how multilingual translation models work and enables us to propose methods to improve performance by training with highly related languages.
Our many-to-one models for high-resource languages and one-to-many models for LRL outperform the best results reported by \citet{aharoni-etal-2019-massively}.%
\end{abstract}

\section{Introduction} \label{sec: intro}


Multilingual training regimens \cite{dong-etal-2015-multi, firat-etal-2016-multi, ha2016toward} are now a key element of natural language processing, especially for low-resource languages (LRLs) \cite{neubig-hu-2018-rapid,aharoni-etal-2019-massively}.
These algorithms are presumed to be helpful because they leverage syntactic or semantic similarities between languages, and transfer processing abilities across language boundaries.

In general, English is used as a central language due to its data availability, and three different multilingual training settings are considered: 
(1) \textit{one-to-many}: training a model with languages pairs from English to many other languages.
(2) \textit{many-to-one}: training a model with languages pairs from many languages to English
(3) \textit{many-to-many}: training a model with the union of the above two settings' data.
(1) and (3) can be used for English to other (En-X) translation, while (2) and (3) can be used for other to English (X-En) translation.

However, multilingual training has not proven equally helpful in every setting.
\citet{arivazhagan2019massively} showed that many-to-one training improves performance over bilingual baselines more than one-to-many does.
In this paper, we consider this result from the point of view of the components of the MT model.
In the many-to-one setting, the model's inputs are from different language distributions so the encoder can be considered a multi-domain model, whereas the decoder is trained on a single distribution.
In the one-to-many setting, it is the opposite: the encoder shares data, and the decoder is multi-domain.
While there are recent studies analyzing multilingual translation models \cite{kudugunta-etal-2019-investigating,voita2020analyzing,aji-etal-2020-neural,mueller-etal-2020-analysis}, in general, they do not (1) examine the impact of different multilingual training settings such as one-to-many and many-to-one, and (2) they do not examine the different components, such as the encoder and the decoder, separately.

This motivates us to ask \emph{``how do various types of multilingual training interact with learning of the encoder and decoder?''}
To answer this question, we set up controlled experiments that decouple the contribution to the encoder and the decoder in various training settings.
We first train multilingual models using many-to-one, one-to-many, or many-to-many training paradigms.
We then compare training bilingual models with and without initializing the encoder or the decoder with parameters learned by multilingual training.
We find that, for LRLs, multilingual training is beneficial to both the encoder and the decoder.
However, surprisingly, for high-resource languages (HRL), we found multilingual training only beneficial to the encoder but not to the decoder.

To further analyze the result, we examine \textit{"to what degree are the learned parameters shared across languages?"}.
We use the head importance estimation method proposed by \citet{michel2019sixteen} as a tool to identify the important attention heads in the model, and measure the consistency between the heads sets that are important for different language pairs. 
The results suggest that the encoder does share parameters across different languages in all settings.
On the other hand, the decoder can treat the representation from the encoder in a language-agnostic way for X-En translation, and less parameter sharing is observed for En-X translation.
Our analyses on parameter sharing also provide a possible explanation to \citet{kudugunta-etal-2019-investigating}'s observation that the representation from the encoder is target-language-dependent.



Our investigation of how multilingual training works leads us to a method for improving MT models. 
With the comprehensive experiments in multilingual settings, for translations in HRL (Ar-En, De-En, He-En, It-En), we discover that fine-tuning multilingual model with target bilingual data outperforms the best results in \citet{aharoni-etal-2019-massively} by 2.99 to 4.63 BLEU score. 
With the analysis of the parameter sharing in the decoder, we are able to identify related languages.
Fine-tuning jointly with the identified related languages boosts low-resource translation (En-Az, En-Be, En-Go, En-Sk) over the best results in \citet{aharoni-etal-2019-massively} by 1.66 to 4.44 BLEU score.
Compared to \citet{neubig-hu-2018-rapid}, our method does not require linguist knowledge, and thus may be more useful for less-studied low-resource languages.

In sum, our contributions are three-fold.
First, our experiments can be used as a diagnostic tool for multilingual translation to investigate how an encoder and a decoder benefit from multilingual training.
Second, our results provide insights into how multilingual translation works.
Third, we improve the translation models based on the findings from our analysis, showing a promising path for future research on multilingual machine translation.


\begin{table}[t]
    \small
    \begin{tabular}{c|c c c c c c c c}
        Lang. & az & be & gl & sk & ar & de & he & it \\
        \midrule
       Size (K) & 6 & 5 & 10 & 61 & 214 & 168 & 212 & 205 \\
    \end{tabular}
    \caption{Training data size.}
    \label{tab:data}
\end{table}

\begin{table*}[h]\centering
\small
\begin{tabular}{cccccccccc}
\toprule
\multicolumn{2}{c}{\multirow{2}{*}{Model}} & \multicolumn{8}{c}{$\rightarrow$ en} \\
\cmidrule{3-10}
 && az & be & gl & sk & ar & de & he & it  \\
 \midrule
 \multicolumn{2}{c}{All-All \cite{aharoni-etal-2019-massively}} & 12.8 & 21.7 & 30.7 & 29.5 & 28.3 & 33.0 & 33.2 & 35.1 \\
 \midrule
 \multicolumn{2}{c}{All-En} & 9.1 & 15.2 & 27.4 & 25.4 & 23.9 & 28.3 & 27.9 & 31.5 \\
 \multicolumn{2}{c}{All-All} & 8.1 & 12.6 & 22.8 & 24.6 & 21.7 & 27.1 & 26.1 & 31.1 \\
 \midrule
\multicolumn{2}{c}{Bilingual Only} & 2.1 & 1.4 & 2.8 & 18.5 & 28.5 & 32.0 & 34.8 & 35.7 \\ 
 \midrule
\multirow{4}{*}{All-En} & Load Enc. & 2.8 & 1.8 & 5.9 & 18.1 & 30.6 & 35.5 & 36.9 & 35.7 \\
 & Load Dec. & 2.5 & 1.8 & 5.7 & 17.8 & 27.2 & 30.3 & 33.2 & 35.7 \\
 & Freeze Enc. & 5.0 & 6.0 & 19.3 & 26.3 & 28.4 & 33.0 & 33.6 & 36.4 \\
 & Freeze Dec. & 3.4 & 4.1 & 16.9 & 24.7 & 28.1 & 31.4 & 33.4 & 33.6 \\
 & Load Both & \textbf{11.5} & 19.0 & 29.9 & 28.00 & 30.4 & 33.1 & 36.2 & 36.7 \\
 \midrule
 \multirow{4}{*}{All-All} & Load Enc. & 5.4 & 7.0 & 20.6 & 28.0 & 30.9 & 35.7 & 37.1 & 38.1 \\
 & Load Dec. & 1.4 & 0.5 & 0.9 & 20.4 & 28.9 & 32.2 & 34.0 & 35.3 \\
 & Freeze Enc. & 3.3 & 5.0 & 9.3 & 23.8 & 25.9 & 32.4 & 32.2 & 34.2 \\
 & Freeze Dec. & 2.0 & 6.2 & 20.1 & 26.9 & 30.1 & 34.4 &35.9 & 36.8 \\
 & Load Both & 11.3 & \textbf{19.4} & \textbf{31.8} & \textbf{29.6} & \textbf{31.3} & \textbf{36.0} & \textbf{37.8} & \textbf{38.7} \\
\bottomrule
\end{tabular}
\caption{Results of translating to English. \textbf{All} in the model name refers to using all 59 languages.}
\label{tab:result_all-en}
\end{table*}
\section{Experimental Settings for Multilingual Training}
Before stepping into our analysis, we first explain our experimental setup.
Following the setting in \citet{aharoni-etal-2019-massively} and \citet{neubig-hu-2018-rapid}, we use the publicly available TED Talks Dataset \cite{qi-etal-2018-pre} is used to train all our machine translation models.
Following \citet{neubig-hu-2018-rapid}, we break words into subwords with BPE jointly learned over all source languages using the \texttt{sentencepiece} toolkit.
The vocabulary size is 32,000.
We perform experiments with the Transformer architecture \cite{vaswani2017attention} using the hyperparameters same as in \cite{arivazhagan2019massively} \footnote{6 layers in both the encoder and the decoder, 8 attention head, state dimension=512,  ffn dimension=2048, label smoothing=0.1}.
All models are implemented and trained using \textit{Fairseq} 0.10.0~\cite{ott2019fairseq}.
We trained multilingual translation models with 60 different languages on the TED Talks Dataset with the three settings described in Section \ref{sec: intro}: \textit{one-to-many}, \textit{many-to-one} and \textit{many-to-many}.
For \textit{one-to-many} and \textit{many-to-many} settings, we add a special language token to the input of the encoder to indicate the target language.
Following \citet{aharoni-etal-2019-massively}, we evaluate our models with BLEU score \cite{papineni-etal-2002-bleu, post-2018-call} on the selected 8 languages.
They are representative of different language families \cite{qi-etal-2018-pre}.
The size of the training is shown in Table~\ref{tab:data}.




\section{How Multilingual Training Benefits Each Component}
\label{sec:how-benefit}
Previous studies have shown that the multilingual training results are generally stronger than the bilingual training \cite{arivazhagan2019massively}.
To understand how multilingual training benefits NMT, we analyze the effect of multilingual training on different components of an NMT model, specifically, the encoder and decoder.

\subsection{Experiments Design}

To study how multilingual training benefits each component, we train models on bilingual data with components initialized differently as follows:
\begin{itemize}
    \item \textbf{Bilingual Only:} Models trained from scratch with no components initialized with parameters learned from multilingual training.
    \item \textbf{Load encoder/decoder:} Models with trainable parameters of either encoder or decoder initialized with parameters learned from multilingual training.
    \item \textbf{Load both:} Models with parameters of both encoder and decoder initialized with parameters learned from multilingual training. This can be seen as fine-tuning the multilingual model on bilingual data.
\end{itemize}

The motivation for this paradigm is that if multilingual training is beneficial to a component, then initializing the parameters of that component should result in improvements over random initialization and training on only bilingual data.
If \emph{load encoder} outperforms \emph{bilingual only}, then we can say that multilingual training is beneficial for the encoder, and if \emph{load decoder} outperforms we can make the analogous conclusion for the decoder.
Thus comparing these models reveals how each component benefit from multilingual training.

We also consider a \textit{load and freeze} setting \citep{thompson-etal-2018-freezing}, where we initialize a component from a multilingual model and freeze its weights when fine-tuning on bilingual data.
For example, in the \textit{load decoder} setting, we train the loaded decoder with a randomly initialized encoder.  
We suspect that learning with randomly initialized component might ruin the other component which is well-trained with multilingual data, especially in the beginning of the training.
Thus, we additionally experiment with this \textit{load and freeze} setting to ensure the multilingual-trained component is not deteriorated.

\begin{table*}[!thp]\centering
\small
\begin{tabular}{cccccccccc}\toprule
\multicolumn{2}{c}{\multirow{2}{*}{Model}} & \multicolumn{8}{c}{en $\rightarrow$} \\\cmidrule{3-10}
 && az & be & gl & sk & ar & de & he & it  \\
 \midrule
 \multicolumn{2}{c}{All-En \cite{aharoni-etal-2019-massively}} & 5.1 & 10.7 & 26.6 & 24.5 & 16.7 & 30.5 & 27.6 & 35.9 \\
 \midrule
\multicolumn{2}{c}{En-All} & 4.9 & 9.0 & 24.2 & 21.9 & 15.1 & 27.9 & 24.1 & 33.3 \\
\multicolumn{2}{c}{All-All} & 3.1 & 6.2 & 20.5 & 18.4 & 12.7 & 24.5 & 21.1 & 30.5 \\
\midrule
 \multicolumn{2}{c}{Bilingual Baseline} &  1.3 & 1.9 & 3.9 & 13.1 & 15.6 & 27.1 & 25.4 & 32.0 \\
 \midrule
 \multirow{4}{*}{En-All} & Load Enc. & 3.0 & 5.6 & 16.7 & 21.7 & \textbf{17.2} & 30.0 & 27.5 & 34.6 \\
 & Load Dec. & 1.3 & 2.0 & 8.1 & 17.4 & 16.0 & 26.7 & 25.8 & 32.6 \\
 & Freeze Enc. & 2.7 & 4.6 & 14.7 & 21.1 & 9.7 & 24.4 & 22.6 & 33.4 \\
 & Freeze Dec. & 1.9 & 3.7 & 14.5 & 17.6 & 16.2 & 28.0 & 25.9 & 33.3 \\
 & Load All & \textbf{6.4} & \textbf{14.7} & \textbf{26.9} & \textbf{23.5} & 17.1 & \textbf{31.1} & \textbf{28.2} & \textbf{34.9} \\
 \midrule
 \multirow{4}{*}{All-All} & Load Enc. & 2.4 & 5.0 & 16.9 & 21.4 & 16.9 & 29.8 & 27.4 & 34.4 \\
 & Load Dec. & 1.1 & 2.2 & 7.0 & 17.5 & 16.0 & 28.1 & 25.6 & 32.5 \\
 & Freeze Enc. & 2.1 & 0.5 & 12.6 & 19.4 & 10.2 & 24.4 & 24.3 &	33.1 \\
  & Freeze Dec. & 0.9 & 4.7 & 15.0 & 18.8 & 15.1 & 27.5 & 24.9 & 32.4 \\
 & Load All & 6.1 & 13.0 & 26.4 & 23.2 & 17.0 & 30.3 & 27.9 & 34.6 \\
\bottomrule
\end{tabular}
\caption{Results of translating from English. \textbf{All} in the model name refers to using all 59 languages.}
\label{tab:result_en-all}
\end{table*}

\subsection{Results and Discussion}

The overall results of X-En and En-X are shown in Table~\ref{tab:result_all-en} and Table~\ref{tab:result_en-all}, respectively.
The difference between the numbers reported in \citet{aharoni-etal-2019-massively} and ours is due to the different batch size and learning rate schedule we use.
In the following section we will discuss the results of our study.
Because they are highly dependent on the training data size (Table~\ref{tab:data}), we discuss the results in two groups: high-resource languages (HRL; referring to \textit{ar}, \textit{de}, \textit{he}, and \textit{it}) and low-resource languages (LRL; referring to \textit{az}, \textit{be}, \textit{gl}, \textit{sk}).%
\footnote{\textit{sk} has intermediate size, and its behavior is not always consistent with the other LRL.}

\subsubsection{Low-Resource Language Results}
For LRLs, we find that multilingual training is generally beneficial to both the encoders and the decoders in all three multilingual models.
Both \textit{load encoder} and \textit{load and freeze decoder} can achieve performance better than the bilingual baseline.
This suggests that the parameters in the encoder and the decoder learned by multilingual training do contain information that is not effectively learned from the smaller bilingual data.

The results also suggest that multilingual training is more beneficial for the encoders than decoders.
In all cases, either \textit{load encoder} or \textit{freeze encoder} outperforms both \textit{load decoder} and \textit{load and freeze decoder}.
However, multilingual training of the encoder and the decoder are complementary; loading both the encoder and the decoder can usually improve the performance over loading only one component.

\subsubsection{High-Resource Language Results}
On HRLs, we find that multilingual training is generally beneficial to the encoders in all three multilingual models, while it is not beneficial for the decoders in some settings.
\textit{Load encoder} consistently outperforms the baseline models, but for the All-En model on X-En translation, and the All-All model on En-X translation, neither \textit{load decoder} nor \textit{load and freeze decoder} outperform the baseline model.

We also observe that multilingual training is generally more beneficial to the encoders than decoders.
In all cases, \textit{load encoder} can achieve performance competitive to \textit{load both} (better or less by within 1 BLEU score).
However, in all cases, both \textit{load decoder} and \textit{load and freeze decoder} have worse performance than \textit{load both}.
Therefore, multilingual training is not as beneficial to the decoders as to the encoders.

\subsection{Discussion}
For LRL, because the size of bilingual training data is small, it is not surprising that multilingual training is beneficial for both the encoder and the decoder.
However, our results are somewhat more surprising for HRL --- it is not trivial that multilingual training is not as beneficial.
In the next section, we focus on explaining the phenomena observed on HRL by investigating how parameters are shared across languages.

\section{How Multilingual Parameters are Shared in Each Component}
\label{sec:how-share}

Given the previous results, we are interested in exactly \emph{how} parameters are shared among different language pairs.
Given that we are using the Transformer architecture, for which multi-head attention is a fundamental component, we use the attention heads as a proxy to analyze how multilingual models work differently when translating between different languages.
Specifically, we analyze our models by identifying the attention heads that are important when translating a language pair.
Measuring the consistency between the sets of important attention heads for two language pairs gives us hints on the extent of parameter sharing. 

\subsection{Head Importance Estimation}
First, we provide some background on head importance estimation, specifically the method proposed by \citet{michel2019sixteen}.

Given a set of multi-head attention modules, each of which can be written as
\begin{equation}
    \text{MHAtt}(x) = 
    \sum_{h=1}^{N_h} \xi_h \text{Att}_{W_q^{(h)}, W_k^{(h)}, W_v^{(h)}}(x),
\end{equation}
where $N_h$ is the number of attention heads, and $\xi_h = 1$ for all $h$. 

The importance of a head can be estimated as 
\begin{equation}
    \tilde{I}_h = \mathbb{E}_{x \sim X} 
    \abs{ \frac{\partial L(x)}{\partial \xi_h} }.
\end{equation}
given a loss function $L$ and input $X$.
Then, the importance score of each head in an attention module is normalized
\begin{equation}
    I_h = \frac{\tilde{I}_h}{\sqrt{\sum_{i}^{N_h} I_h^2}}.
    \label{eq:head-score}
\end{equation}
Note that when the input $X$ is different, the estimated importance score can be different.
Therefore, when different language pairs are fed in, the important heads identified can be different.
We denote the set of attention head scores estimated on translation from language $l_a$ to language $l_b$ as $H(l_a, l_b)$.
We denote the scores of attention heads in a component by using superscript.
For example, $H^{enc}$ represents the scores of the heads in a encoder.

\subsection{Measuring Parameter Sharing by Correlation of Head Scores}
With the attention head importance scores estimated by Equation~\ref{eq:head-score}, we can investigate how parameters are shared across languages.
For each of the En-All, All-En, All-All multilingual models, we estimated a set of head-importance scores $H(l_a, l_b)$ for each language pair $(l_a, l_b)$ in the training setting.
We calculate the head scores with the training loss function (MLE with label smoothing) and 100K randomly sampled sentences in the training set.

To investigate how much parameters are shared by two pairs of languages $(l_a, l_b)$ and $(l_c, l_d)$, we measure the agreement between $H(l_a, l_b)$ and $H(l_c, l_d)$.
If a head is important for both of $(l_a, l_b)$ and $(l_c, l_d)$, then important parameters for translating are shared.
Thus high agreement suggests high parameter sharing.

To quantify the agreement between two score sets, we use Spearman's rank correlation \cite{spearman1987proof}.
A rank-based correlation metric is used because the importance estimation was originally proposed to order attention heads in a model.
Higher correlation implies higher agreement and thus implies higher parameter sharing.
For each of the En-All, All-En, All-All models, we calculate the correlation between $H(l_a, l_b)$ and $H(l_c, l_d)$ for all language pairs $(l_a, l_b)$ and $(l_c, l_d)$ that are used to train the model.
The detailed correlation computation process can be found in Appendix~\ref{app:correlation}.
We plot the correlation matrices of the head scores (included in appendix) and summarize them in Table~\ref{tab:corr}.
We also compare the top-10 most important heads for every language pairs with F1 scores, and observe similar results.
We include the statistics in appendix.

\begin{table}[]
    \centering
    \begin{tabular}{c  c  c  c }
        \toprule
        Model   & Lang. Pair &  $H^{enc}$ & $H^{dec}$ \\
        \midrule
        All-En  &  X-En & .871 (.086) & .973 (.023) \\
        En-All  &  En-X & .806 (.153) & .720 (.150) \\
        All-All &  X-En & .898 (.073) & .967 (.029) \\
        All-All &  En-X & .813 (.126) & .762 (.141) \\
        \bottomrule
    \end{tabular}
    \caption{Correlation between the attention head scores when estimated using different language pairs.}
    \label{tab:corr}
\end{table}

\subsection{How Multilingual Translation Models Share}
Results in Table~\ref{tab:corr} combined with Section~\ref{sec:how-benefit} provides the insights into how multilingual translation models work with respect to cross-lingual sharing:


\paragraph{Encoder for En-X:}

It is natural that the encoder from En-X likely benefit from multilingual training because it can generate representations tailored for different target languages with shared parameters.
En-X is a set of language pairs where the source language is always English.
Therefore, if the prepended target language token is ignored, the inputs of the encoders for all pairs in En-X are from one identical distribution.
This is in contrast to X-En pairs, where the inputs are in different languages.
However, for the encoders, we observe from Table~\ref{tab:corr} that the average correlation scores of En-X pairs (0.806 and 0.813), are lower than the correlation scores of X-En pairs (0.871 and 0.898). 
\citeauthor{kudugunta-etal-2019-investigating} discovers that the representation of the encoder is target-language-dependent. 
Thus we conjecture that some parameters may be used to generate representation tailored for the target languages.
At the same time, since the inputs are from a single distribution (English) for different target languages, a large portion of parameters may still be shareable across target languages.
Therefore, in this case, multilingual training is beneficial.

\paragraph{Encoder for X-En:}
For X-En language pairs, the input of the encoder is multilingual, which means the input from different X-En language pairs has distinct distribution.
However, the correlation between different source languages is still high.
It shows that high parameters sharing in the encoder is possible.

\paragraph{Decoder for En-X:}
The decoders for En-X have the lowest correlation. 
From the correlation matrix, we do see some parameter sharing between some language pairs.
However, larger model capacity might be required for a model to be proficient in all the languages.

\paragraph{Decoder for X-En:}
The decoder have average correlation as high as 0.973 and 0.967 for All-En and All-All models respectively.
This suggests that to decode intermediate representation encoded by the encoder, the decoder use almost the same set of parameters.
However, \citeauthor{kudugunta-etal-2019-investigating} shows that the representation encoded by the encoder is not language-agnostic.
A possible explanation is that the important parameters of the decoder are highly determined by the target output, which is always in English.
Therefore, even though the encoder representation is not language-agnostic, it is still difficult to learn parameters reflecting the difference.
It suggests why multilingual training does not benefit the decoder in the X-En setting.
The set of English sentences is almost the same for all the HRL pairs in the TED Talks dataset, so multilingual training can hardly provide more unique English sentences than bilingual training does.
If the decoder is dedicated for generation, multilingual training cannot expose the decoder to more diverse data.
Therefore the multilingually trained decoder does not perform better than the bilingual one.

\begin{table*}
    \center
    \begin{tabular}{lccccccccc}
        \toprule
        Model    & az & be & gl & sk & ar & de & he & it \\
        \midrule
        All-All & 8.1 & 12.6 & 22.8 & 24.6 & 21.7 & 27.1 & 26.1 & 31.1 \\
        + f.t. on All-En & 10.5 & 17.5 & 29.7 & 28.1 & 25.9 & 31.3 & 30.5 & 34.0 \\
        + f.t. on All-En \& related & 10.5 & 17.4 & 28.3 & 27.0 & 25.1 & 30.0 & 29.9 & 32.7 \\
        \bottomrule
    \end{tabular}
    \caption{Performance of All-All model fine-tuned on All-En pairs and fine-tuned on the union of All-En pairs and related En-All languages.}
    \label{tab:related}
\end{table*}

\begin{table*}[htp!]
    \center
    \begin{tabular}{lcccccccc}
        \toprule
        Model    & az & be & gl & sk & ar & de & he & it \\
        \midrule
        En-All \cite{aharoni-etal-2019-massively} & 5.1 & 10.7 & 26.6 & 24.5 & 16.7 & 30.5 & 27.6 & 35.9 \\
        \midrule
        Bilingual Baseline &  1.3 & 1.9 & 3.9 & 13.1 & 15.6 & 27.1 & 25.4 & 32.0 \\
        \hline
        All-All & 3.1 & 6.2 & 20.5 & 18.4 & 12.7 & 24.5 & 21.1 & 30.5 \\
        All-All w/ f.t. on related clusters & 7.9 & 12.8 & 27.5 & 24.9 & - & 30.2 & 27.0 & 35.4 \\
        All-All w/ f.t. on random groups & 6.9 & 13.3 & 22.5 & 24.3 & - & - & 27.5 & 35.2 \\
        \hline
        En-All & 4.9 & 9.00 & 24.2 & 21.9 & 15.1 & 27.9 & 24.1 & 33.3 \\
        En-All w/ f.t. on related clusters & \textbf{7.9} & 13.9 & 21.0 & \textbf{26.2} & 16.7 & 30.4 & 27.1 & 35.4 \\
        En-All w/ f.t. on random groups & 7.0 & 13.1 & 23.1 & 24.7 & - & - & 27.6 & 35.2 \\
        \hline
         Load En-All w/ f.t. on closest & 7.8 & \textbf{15.2} & \textbf{28.6} &  & & & & \\
        \bottomrule
    \end{tabular}
    \caption{Performance of En-All model without and with fine-tuning on language clusters.}
    \label{tab:cluster}
\end{table*}

\section{Improving Translation Based on the Degree of Parameter Sharing}
\label{sec:improve}

Insights from the previous section provide us with a new way to choose languages for multilingual training. 
In previous work \citep{lin-etal-2019-choosing,oncevay-etal-2020-bridging}, choosing on languages with similar linguistic properties is a popular practice.
However, \citet{mueller-etal-2020-analysis} found the effect is highly language-dependent. 
Sometimes training with similar languages might be worse than training on a set of unrelated languages.
Here we otherwise propose an entirely model-driven way to find related languages to improve multilingual translation models.
We explore choosing languages where parameters can be better shared.

\subsection{Improving X-En by Related En-X Pairs}

In the All-All model, we notice low parameter sharing between En-X and X-En pairs.
The average correlation between $H^{enc}(En, X)$ and $H^{enc}(X, En)$ is 0.44 (std: 0.17).
The average correlation between $H^{dec}(En, X)$ and $H^{dec}(X, En)$ is 0.49 (std: 0.13).
It provides a possible explanation why training with both the En-X and the X-En pairs only brings little improvement over training with only En-X alone or with X-En alone.

The low correlation combined with results in Section~\ref{sec:how-benefit} motivate us to experiment on improving X-En with related En-X pairs.
Section~\ref{sec:how-benefit} shows that the multilingual decoder has less advantage than the encoder. 
This may suggest the inefficiency of parameter sharing in the decoder.
Therefore we experiment on choosing a set of related languages based on the degree of parameter in the decoder.
We choose the language set $L$ such that for all $l \in L$, the average correlation 
$\frac{1}{60} \sum_{l_i=1}^{60} \mathrm{Corr}(H^{dec}(En, l), H^{dec}(l_i, En))$
is higher than 0.60. 

Results are shown in Table~\ref{tab:related}.
Even though fine-tuning on related languages improves the overall performance, it is not better than fine-tuning on the All-En pairs only.
Also, the average correlation between $H^{dec}(En, l_a)$ and $H^{dec}(l_b, En)$ is not improved.
Our experiment demonstrates the difficulty of sharing parameters between All-En pairs and En-All pairs.
We leave this problem for future work.

\subsection{Improving En-X by Language Clusters}
\label{sec:lang-cluster}

The low correlation between attention head scores of language pairs motivates us to improve the performance of En-X using related language pairs. 
As shown in Table~\ref{tab:corr}, the decoders have the lowest correlation scores.
We conjecture that it is due to the difficulty of sharing parameters between distant languages.
Thus, we seek for finding related language sets, in each of which parameters can be shared.

\begin{figure}
    \centering
    \includegraphics[width=0.85\linewidth]{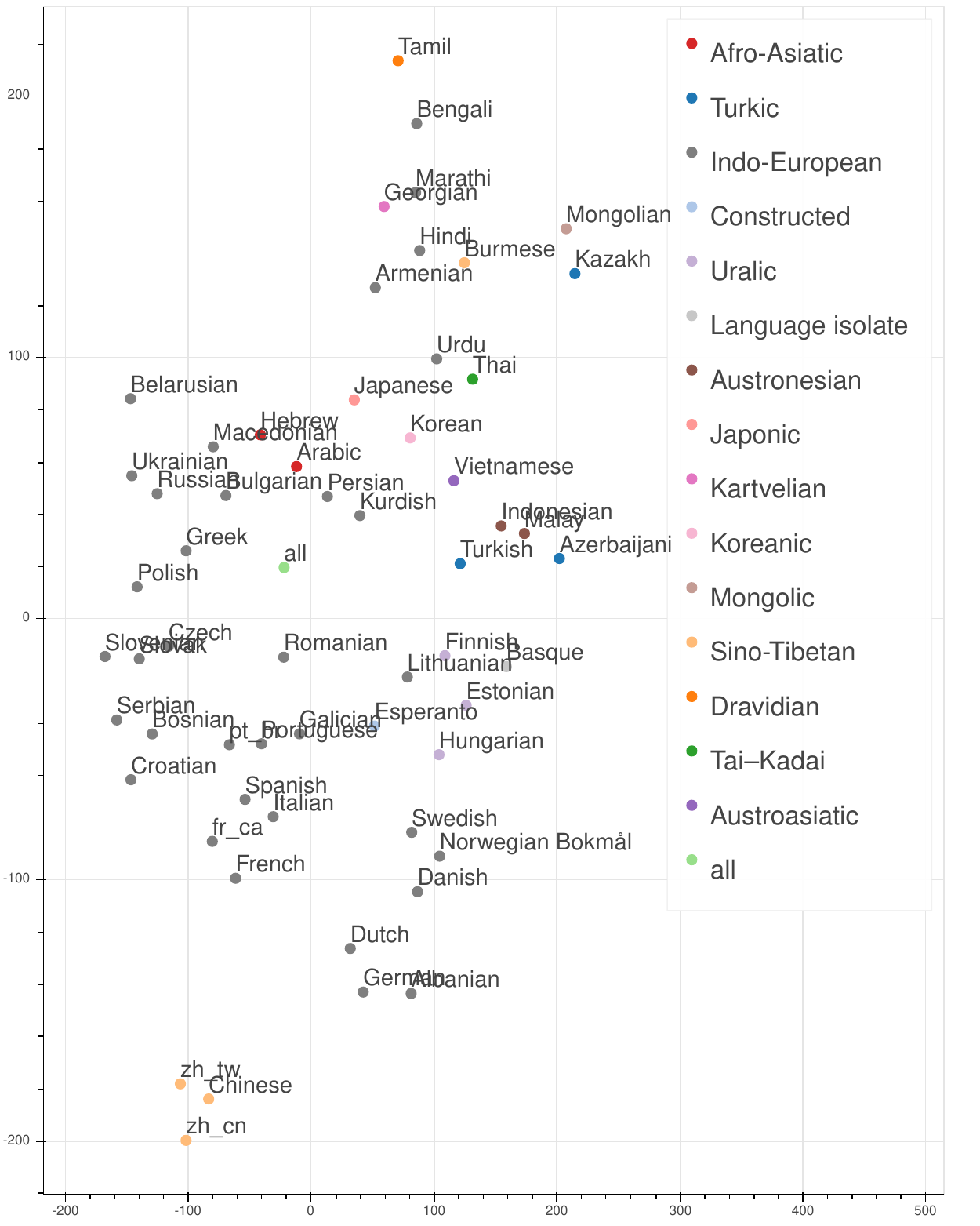}
    \caption{Visualization of the En-All decoder head scores of languages by t-SNE.}
    \label{fig:tsne}
\end{figure}

Again, we resort to the attention head importance scores to find the related languages.
Our intuition is that related languages would share many parameters in between and training a model on related languages would be helpful.
As a sanity check of our idea, we first use t-SNE~\cite{maaten2008visualizing} to reduce the dimension of head-importance scores $H(l_a, l_b)$. 
We only focus on heads in the decoders, because the correlation score between $H_{(\mathrm{En}, l_c)}$ and $H_{(\mathrm{En}, l_d)}$ is lower in average for the decoders.
The result visualized in Figure~\ref{fig:tsne} illustrates that, 
the distance between $H_{(\mathrm{En}, l_c)}$ and $H_{(\mathrm{En}, l_d)}$ tend to be shorter if languages $l_c$ and $l_d$ are linguistically related.
Hence, determining related languages with head score $H_{(\mathrm{En}, l)}$ should be reasonable.

We then fine-tune multilingual models on related language clusters.
Related languages clusters are determined by k-mean++~\cite{arthur2006k} with $k=5$.
We consider clusters that cover all of the four low-resource languages.
For the All-All model, one of the cluster we consider contains Be, Gl, De, He, It, and the other one contains Az.
For the En-All model, we also experiment with two clusters. 
One includes Ar, De, He, It, and the other includes Az, Be, Gl, Sk.
As a baseline, we also experiment with random groups.
They are groups generated by randomly splitting the 59 target languages.

The results are shown in Table~\ref{tab:cluster}.
For both the En-All and the All-All model, except En-Gl, fine-tuning on clusters can improve performance on all the considered language pairs consistently.
For LRLs, fine-tuning on related language clusters is also better than fine-tuning on random groups in general. 
To verify whether this improvement is brought by increased parameter sharing in the decoders, we check the correlation between $H^{dec}$ after fine-tuning.
The results shown in Table~\ref{tab:corr-cluster} shows improvements after fine-tuning on the clusters.

For low-resource language pairs En-Az, En-Be, En-Sk on the En-All model, we notice that only few languages are highly correlated with them (with correlation $>$ 0.80).
Therefore, we also experiment with fine-tuning the En-All model with only the language pairs with high correlation scores ($> 0.80$) for each of the three pairs
, which boosts the performance of En-Be to 15.2 and En-Sk to 28.6. 

\begin{table}[]
    \centering
    \begin{tabular}{c  c  c  c }
        \toprule
        Model   & $H^{dec}$ w/o f.t. & $H^{dec}$ w/ f.t. \\
        \midrule
        All-All &  .762 (.141) & .894 (.069) \\
        En-All (HL) &  .855 (.066) &  .866 (.065) \\
        En-All (LL) &  .826 (.096) & .834 (.091) \\
        \bottomrule
    \end{tabular}
    \caption{Correlation between the decoder attention head scores when estimated using the language pairs in the cluster. HL and LL represent the cluster that includes HRL and the one that includes LRL respectively.}
    \label{tab:corr-cluster}
\end{table}

\section{Related Work}

The early attempts of multilingual training for machine translation use a single model to translate between multiple languages \cite{dong-etal-2015-multi, firat-etal-2016-multi, ha2016toward}.
Those works find multilingual NMT models are appealing because they not only give us a simple paradigm to handle mapping between multiple languages, but also improve performance on low and zero-resource languages pairs \cite{gu-etal-2018-universal}.
However, how multilingual training contributes to components in the translation model still remains unknown.

There are some attempts at analyzing and explaining the translation models.
\citet{thompson-etal-2018-freezing} analyze the contribution of different components of NMT model to domain adaptation by freezing the weights of components during continued training.
\citet{arivazhagan2019massively} provide an comprehensive study on the state-of-the-art multilingual NMT model in different training and testing scenarios.
\citet{sachan-neubig-2018-parameter} experiment with different parameter sharing strategies in Transformer models, showing that sharing parameters of embedding, key and query performs well for \textit{one-to-many} settings.
\citet{artetxe-etal-2020-cross} shows the strong transferability of monolingual representation to different languages.
The intermediate representation of BERT can be language-agnostic if we freeze the embeddings during training.
The deficiency of the \textit{one-to-many} setting is explored in \cite{johnson-etal-2017-googles}.
They find only the \textit{many-to-one} setting consistently improves the performance across languages.
\citet{wang-etal-2018-three} also explore problems of the \textit{one-to-many} setting, and show language-specific components are effective to improve the performance.
\citet{voita2020analyzing} analyzes how generated sentences of NMT models are influenced by context in the encoder and decoder.
The attempt to investigate encoder and decoder separately is similar to our work.
\citet{rothe-etal-2020-leveraging} explores how pretrained checkpoints can benefit the encoder and the decoder in a translation model.  
\citet{zhang2021share} investigate the trade-off between language-specific and shared capacity of layers in a multilingual NMT model.

Multi-head attention has been shown effective in different NLP tasks.
Beyond improving performance, multi-head attention can help with subject-verb agreement \cite{tang-etal-2018-self}, and some heads are predictive of dependency structures \cite{raganato-tiedemann-2018-analysis}.
\citet{htut2019attention} and \citet{clark2019does} report that heads in BERT attend significantly more to words in certain syntactic position.
They show some heads seem to specialize in certain types of syntactic relations.
\citet{michel2019sixteen},  \citet{voita-etal-2019-analyzing}, and \citet{behnke-heafield-2020-losing} study the importance of different attention heads in NMT models, and suggest that we can prune those attention heads which are less important.
\citet{brix-etal-2020-successfully} also shows pruning NMT models can improve the sparsity level to optimize the memory usage and inference speed.

However, all previous works do not directly investigate how encoder and decoder of NMT models benefit from multilingual training, which is the key question of why multilingual training works.
To our best knowledge, we are the first to tackle the question, and our analysis can be used to further improve multilingual NMT models.

\section{Conclusion}
In this work, we have the following findings:
1) 
In Section~\ref{sec:how-benefit}, we examine how multilingual training contributes to each of the components in a machine translation model. 
We discover that, while multilingual training is beneficial to the encoders, it is less beneficial to the decoders.
2)
In Section~\ref{sec:how-share}, our analysis of important attention heads provides insight into the behavior of multilingual components.
Results suggest that the encoder in the En-All model may generate target-language-specific representation, while the behavior of the decoder of the All-En model may be source-language-agnostic.
In addition, in the All-All model, we observe indications of lower parameter sharing between X-En pairs and En-X pairs. 
3)
In Section~\ref{sec:improve}, we explore approaches to improve the model based on our findings.
On En-X translation, we outperform the best results in \cite{aharoni-etal-2019-massively}.
With our proposed analysis as diagnostic tools, future work may further improve the multilingual systems.

\bibliography{anthology,custom}

\begin{thebibliography}{36}
\expandafter\ifx\csname natexlab\endcsname\relax\def\natexlab#1{#1}\fi

\bibitem[{Aharoni et~al.(2019)Aharoni, Johnson, and
  Firat}]{aharoni-etal-2019-massively}
Roee Aharoni, Melvin Johnson, and Orhan Firat. 2019.
\newblock \href {https://doi.org/10.18653/v1/N19-1388} {Massively multilingual
  neural machine translation}.
\newblock In \emph{Proceedings of the 2019 Conference of the North {A}merican
  Chapter of the Association for Computational Linguistics: Human Language
  Technologies, Volume 1 (Long and Short Papers)}, pages 3874--3884,
  Minneapolis, Minnesota. Association for Computational Linguistics.

\bibitem[{Aji et~al.(2020)Aji, Bogoychev, Heafield, and
  Sennrich}]{aji-etal-2020-neural}
Alham~Fikri Aji, Nikolay Bogoychev, Kenneth Heafield, and Rico Sennrich. 2020.
\newblock \href {https://doi.org/10.18653/v1/2020.acl-main.688} {In neural
  machine translation, what does transfer learning transfer?}
\newblock In \emph{Proceedings of the 58th Annual Meeting of the Association
  for Computational Linguistics}, pages 7701--7710, Online. Association for
  Computational Linguistics.

\bibitem[{Arivazhagan et~al.(2019)Arivazhagan, Bapna, Firat, Lepikhin, Johnson,
  Krikun, Chen, Cao, Foster, Cherry et~al.}]{arivazhagan2019massively}
Naveen Arivazhagan, Ankur Bapna, Orhan Firat, Dmitry Lepikhin, Melvin Johnson,
  Maxim Krikun, Mia~Xu Chen, Yuan Cao, George Foster, Colin Cherry, et~al.
  2019.
\newblock Massively multilingual neural machine translation in the wild:
  Findings and challenges.
\newblock \emph{arXiv preprint arXiv:1907.05019}.

\bibitem[{Artetxe et~al.(2020)Artetxe, Ruder, and
  Yogatama}]{artetxe-etal-2020-cross}
Mikel Artetxe, Sebastian Ruder, and Dani Yogatama. 2020.
\newblock \href {https://doi.org/10.18653/v1/2020.acl-main.421} {On the
  cross-lingual transferability of monolingual representations}.
\newblock In \emph{Proceedings of the 58th Annual Meeting of the Association
  for Computational Linguistics}, pages 4623--4637, Online. Association for
  Computational Linguistics.

\bibitem[{Arthur and Vassilvitskii(2007)}]{arthur2006k}
David Arthur and Sergei Vassilvitskii. 2007.
\newblock K-means++: The advantages of careful seeding.
\newblock In \emph{Proceedings of the Eighteenth Annual ACM-SIAM Symposium on
  Discrete Algorithms}, SODA '07, page 1027–1035, USA. Society for Industrial
  and Applied Mathematics.

\bibitem[{Behnke and Heafield(2020)}]{behnke-heafield-2020-losing}
Maximiliana Behnke and Kenneth Heafield. 2020.
\newblock \href {https://doi.org/10.18653/v1/2020.emnlp-main.211} {Losing heads
  in the lottery: Pruning transformer attention in neural machine translation}.
\newblock In \emph{Proceedings of the 2020 Conference on Empirical Methods in
  Natural Language Processing (EMNLP)}, pages 2664--2674, Online. Association
  for Computational Linguistics.

\bibitem[{Brix et~al.(2020)Brix, Bahar, and Ney}]{brix-etal-2020-successfully}
Christopher Brix, Parnia Bahar, and Hermann Ney. 2020.
\newblock \href {https://doi.org/10.18653/v1/2020.acl-main.360} {Successfully
  applying the stabilized lottery ticket hypothesis to the transformer
  architecture}.
\newblock In \emph{Proceedings of the 58th Annual Meeting of the Association
  for Computational Linguistics}, pages 3909--3915, Online. Association for
  Computational Linguistics.

\bibitem[{Clark et~al.(2019)Clark, Khandelwal, Levy, and
  Manning}]{clark2019does}
Kevin Clark, Urvashi Khandelwal, Omer Levy, and Christopher~D Manning. 2019.
\newblock What does bert look at? an analysis of bert's attention.
\newblock In \emph{Proceedings of the 2019 ACL Workshop BlackboxNLP: Analyzing
  and Interpreting Neural Networks for NLP}.

\bibitem[{Dong et~al.(2015)Dong, Wu, He, Yu, and Wang}]{dong-etal-2015-multi}
Daxiang Dong, Hua Wu, Wei He, Dianhai Yu, and Haifeng Wang. 2015.
\newblock \href {https://doi.org/10.3115/v1/P15-1166} {Multi-task learning for
  multiple language translation}.
\newblock In \emph{Proceedings of the 53rd Annual Meeting of the Association
  for Computational Linguistics and the 7th International Joint Conference on
  Natural Language Processing (Volume 1: Long Papers)}, pages 1723--1732,
  Beijing, China. Association for Computational Linguistics.

\bibitem[{Firat et~al.(2016)Firat, Cho, and Bengio}]{firat-etal-2016-multi}
Orhan Firat, Kyunghyun Cho, and Yoshua Bengio. 2016.
\newblock \href {https://doi.org/10.18653/v1/N16-1101} {Multi-way, multilingual
  neural machine translation with a shared attention mechanism}.
\newblock In \emph{Proceedings of the 2016 Conference of the North {A}merican
  Chapter of the Association for Computational Linguistics: Human Language
  Technologies}, pages 866--875, San Diego, California. Association for
  Computational Linguistics.

\bibitem[{Gu et~al.(2018)Gu, Hassan, Devlin, and Li}]{gu-etal-2018-universal}
Jiatao Gu, Hany Hassan, Jacob Devlin, and Victor~O.K. Li. 2018.
\newblock \href {https://doi.org/10.18653/v1/N18-1032} {Universal neural
  machine translation for extremely low resource languages}.
\newblock In \emph{Proceedings of the 2018 Conference of the North {A}merican
  Chapter of the Association for Computational Linguistics: Human Language
  Technologies, Volume 1 (Long Papers)}, pages 344--354, New Orleans,
  Louisiana. Association for Computational Linguistics.

\bibitem[{Ha et~al.(2016)Ha, Niehues, and Waibel}]{ha2016toward}
Thanh-Le Ha, Jan Niehues, and Alexander Waibel. 2016.
\newblock Toward multilingual neural machine translation with universal encoder
  and decoder.
\newblock \emph{arXiv preprint arXiv:1611.04798}.

\bibitem[{Htut et~al.(2019)Htut, Phang, Bordia, and Bowman}]{htut2019attention}
Phu~Mon Htut, Jason Phang, Shikha Bordia, and Samuel~R Bowman. 2019.
\newblock Do attention heads in bert track syntactic dependencies?
\newblock \emph{arXiv preprint arXiv:1911.12246}.

\bibitem[{Johnson et~al.(2017)Johnson, Schuster, Le, Krikun, Wu, Chen, Thorat,
  Vi{\'e}gas, Wattenberg, Corrado, Hughes, and
  Dean}]{johnson-etal-2017-googles}
Melvin Johnson, Mike Schuster, Quoc~V. Le, Maxim Krikun, Yonghui Wu, Zhifeng
  Chen, Nikhil Thorat, Fernanda Vi{\'e}gas, Martin Wattenberg, Greg Corrado,
  Macduff Hughes, and Jeffrey Dean. 2017.
\newblock \href {https://doi.org/10.1162/tacl_a_00065} {{G}oogle{'}s
  multilingual neural machine translation system: Enabling zero-shot
  translation}.
\newblock \emph{Transactions of the Association for Computational Linguistics},
  5:339--351.

\bibitem[{Kudugunta et~al.(2019)Kudugunta, Bapna, Caswell, and
  Firat}]{kudugunta-etal-2019-investigating}
Sneha Kudugunta, Ankur Bapna, Isaac Caswell, and Orhan Firat. 2019.
\newblock \href {https://doi.org/10.18653/v1/D19-1167} {Investigating
  multilingual {NMT} representations at scale}.
\newblock In \emph{Proceedings of the 2019 Conference on Empirical Methods in
  Natural Language Processing and the 9th International Joint Conference on
  Natural Language Processing (EMNLP-IJCNLP)}, pages 1565--1575, Hong Kong,
  China. Association for Computational Linguistics.

\bibitem[{Lin et~al.(2019)Lin, Chen, Lee, Li, Zhang, Xia, Rijhwani, He, Zhang,
  Ma, Anastasopoulos, Littell, and Neubig}]{lin-etal-2019-choosing}
Yu-Hsiang Lin, Chian-Yu Chen, Jean Lee, Zirui Li, Yuyan Zhang, Mengzhou Xia,
  Shruti Rijhwani, Junxian He, Zhisong Zhang, Xuezhe Ma, Antonios
  Anastasopoulos, Patrick Littell, and Graham Neubig. 2019.
\newblock \href {https://doi.org/10.18653/v1/P19-1301} {Choosing transfer
  languages for cross-lingual learning}.
\newblock In \emph{Proceedings of the 57th Annual Meeting of the Association
  for Computational Linguistics}, pages 3125--3135, Florence, Italy.
  Association for Computational Linguistics.

\bibitem[{Maaten and Hinton(2008)}]{maaten2008visualizing}
Laurens van~der Maaten and Geoffrey Hinton. 2008.
\newblock Visualizing data using t-sne.
\newblock \emph{Journal of machine learning research}, 9(Nov):2579--2605.

\bibitem[{Michel et~al.(2019)Michel, Levy, and Neubig}]{michel2019sixteen}
Paul Michel, Omer Levy, and Graham Neubig. 2019.
\newblock Are sixteen heads really better than one?
\newblock In \emph{Advances in Neural Information Processing Systems}, pages
  14014--14024.

\bibitem[{Mueller et~al.(2020)Mueller, Nicolai, McCarthy, Lewis, Wu, and
  Yarowsky}]{mueller-etal-2020-analysis}
Aaron Mueller, Garrett Nicolai, Arya~D. McCarthy, Dylan Lewis, Winston Wu, and
  David Yarowsky. 2020.
\newblock \href {https://www.aclweb.org/anthology/2020.lrec-1.458} {An analysis
  of massively multilingual neural machine translation for low-resource
  languages}.
\newblock In \emph{Proceedings of the 12th Language Resources and Evaluation
  Conference}, pages 3710--3718, Marseille, France. European Language Resources
  Association.

\bibitem[{Neubig and Hu(2018)}]{neubig-hu-2018-rapid}
Graham Neubig and Junjie Hu. 2018.
\newblock \href {https://doi.org/10.18653/v1/D18-1103} {Rapid adaptation of
  neural machine translation to new languages}.
\newblock In \emph{Proceedings of the 2018 Conference on Empirical Methods in
  Natural Language Processing}, pages 875--880, Brussels, Belgium. Association
  for Computational Linguistics.

\bibitem[{Oncevay et~al.(2020)Oncevay, Haddow, and
  Birch}]{oncevay-etal-2020-bridging}
Arturo Oncevay, Barry Haddow, and Alexandra Birch. 2020.
\newblock \href {https://doi.org/10.18653/v1/2020.emnlp-main.187} {Bridging
  linguistic typology and multilingual machine translation with multi-view
  language representations}.
\newblock In \emph{Proceedings of the 2020 Conference on Empirical Methods in
  Natural Language Processing (EMNLP)}, pages 2391--2406, Online. Association
  for Computational Linguistics.

\bibitem[{Ott et~al.(2019)Ott, Edunov, Baevski, Fan, Gross, Ng, Grangier, and
  Auli}]{ott2019fairseq}
Myle Ott, Sergey Edunov, Alexei Baevski, Angela Fan, Sam Gross, Nathan Ng,
  David Grangier, and Michael Auli. 2019.
\newblock fairseq: A fast, extensible toolkit for sequence modeling.
\newblock In \emph{Proceedings of NAACL-HLT 2019: Demonstrations}.

\bibitem[{Papineni et~al.(2002)Papineni, Roukos, Ward, and
  Zhu}]{papineni-etal-2002-bleu}
Kishore Papineni, Salim Roukos, Todd Ward, and Wei-Jing Zhu. 2002.
\newblock \href {https://doi.org/10.3115/1073083.1073135} {{B}leu: a method for
  automatic evaluation of machine translation}.
\newblock In \emph{Proceedings of the 40th Annual Meeting of the Association
  for Computational Linguistics}, pages 311--318, Philadelphia, Pennsylvania,
  USA. Association for Computational Linguistics.

\bibitem[{Post(2018)}]{post-2018-call}
Matt Post. 2018.
\newblock \href {https://doi.org/10.18653/v1/W18-6319} {A call for clarity in
  reporting {BLEU} scores}.
\newblock In \emph{Proceedings of the Third Conference on Machine Translation:
  Research Papers}, pages 186--191, Brussels, Belgium. Association for
  Computational Linguistics.

\bibitem[{Qi et~al.(2018)Qi, Sachan, Felix, Padmanabhan, and
  Neubig}]{qi-etal-2018-pre}
Ye~Qi, Devendra Sachan, Matthieu Felix, Sarguna Padmanabhan, and Graham Neubig.
  2018.
\newblock \href {https://doi.org/10.18653/v1/N18-2084} {When and why are
  pre-trained word embeddings useful for neural machine translation?}
\newblock In \emph{Proceedings of the 2018 Conference of the North {A}merican
  Chapter of the Association for Computational Linguistics: Human Language
  Technologies, Volume 2 (Short Papers)}, pages 529--535, New Orleans,
  Louisiana. Association for Computational Linguistics.

\bibitem[{Raganato and Tiedemann(2018)}]{raganato-tiedemann-2018-analysis}
Alessandro Raganato and J{\"o}rg Tiedemann. 2018.
\newblock \href {https://doi.org/10.18653/v1/W18-5431} {An analysis of encoder
  representations in transformer-based machine translation}.
\newblock In \emph{Proceedings of the 2018 {EMNLP} Workshop {B}lackbox{NLP}:
  Analyzing and Interpreting Neural Networks for {NLP}}, pages 287--297,
  Brussels, Belgium. Association for Computational Linguistics.

\bibitem[{Rothe et~al.(2020)Rothe, Narayan, and
  Severyn}]{rothe-etal-2020-leveraging}
Sascha Rothe, Shashi Narayan, and Aliaksei Severyn. 2020.
\newblock \href {https://doi.org/10.1162/tacl_a_00313} {Leveraging pre-trained
  checkpoints for sequence generation tasks}.
\newblock \emph{Transactions of the Association for Computational Linguistics},
  8:264--280.

\bibitem[{Sachan and Neubig(2018)}]{sachan-neubig-2018-parameter}
Devendra Sachan and Graham Neubig. 2018.
\newblock \href {https://doi.org/10.18653/v1/W18-6327} {Parameter sharing
  methods for multilingual self-attentional translation models}.
\newblock In \emph{Proceedings of the Third Conference on Machine Translation:
  Research Papers}, pages 261--271, Brussels, Belgium. Association for
  Computational Linguistics.

\bibitem[{Spearman(1987)}]{spearman1987proof}
Charles Spearman. 1987.
\newblock The proof and measurement of association between two things.
\newblock \emph{The American journal of psychology}, 100(3/4):441--471.

\bibitem[{Tang et~al.(2018)Tang, M{\"u}ller, Rios, and
  Sennrich}]{tang-etal-2018-self}
Gongbo Tang, Mathias M{\"u}ller, Annette Rios, and Rico Sennrich. 2018.
\newblock \href {https://doi.org/10.18653/v1/D18-1458} {Why self-attention? a
  targeted evaluation of neural machine translation architectures}.
\newblock In \emph{Proceedings of the 2018 Conference on Empirical Methods in
  Natural Language Processing}, pages 4263--4272, Brussels, Belgium.
  Association for Computational Linguistics.

\bibitem[{Thompson et~al.(2018)Thompson, Khayrallah, Anastasopoulos, McCarthy,
  Duh, Marvin, McNamee, Gwinnup, Anderson, and
  Koehn}]{thompson-etal-2018-freezing}
Brian Thompson, Huda Khayrallah, Antonios Anastasopoulos, Arya~D. McCarthy,
  Kevin Duh, Rebecca Marvin, Paul McNamee, Jeremy Gwinnup, Tim Anderson, and
  Philipp Koehn. 2018.
\newblock \href {https://doi.org/10.18653/v1/W18-6313} {Freezing subnetworks to
  analyze domain adaptation in neural machine translation}.
\newblock In \emph{Proceedings of the Third Conference on Machine Translation:
  Research Papers}, pages 124--132, Brussels, Belgium. Association for
  Computational Linguistics.

\bibitem[{Vaswani et~al.(2017)Vaswani, Shazeer, Parmar, Uszkoreit, Jones,
  Gomez, Kaiser, and Polosukhin}]{vaswani2017attention}
Ashish Vaswani, Noam Shazeer, Niki Parmar, Jakob Uszkoreit, Llion Jones,
  Aidan~N Gomez, {\L}ukasz Kaiser, and Illia Polosukhin. 2017.
\newblock Attention is all you need.
\newblock \emph{Advances in neural information processing systems},
  30:5998--6008.

\bibitem[{Voita et~al.(2019{\natexlab{a}})Voita, Talbot, Moiseev, Sennrich, and
  Titov}]{voita2020analyzing}
Elena Voita, David Talbot, Fedor Moiseev, Rico Sennrich, and Ivan Titov.
  2019{\natexlab{a}}.
\newblock \href {https://doi.org/10.18653/v1/P19-1580} {Analyzing multi-head
  self-attention: Specialized heads do the heavy lifting, the rest can be
  pruned}.
\newblock In \emph{Proceedings of the 57th Annual Meeting of the Association
  for Computational Linguistics}, pages 5797--5808, Florence, Italy.
  Association for Computational Linguistics.

\bibitem[{Voita et~al.(2019{\natexlab{b}})Voita, Talbot, Moiseev, Sennrich, and
  Titov}]{voita-etal-2019-analyzing}
Elena Voita, David Talbot, Fedor Moiseev, Rico Sennrich, and Ivan Titov.
  2019{\natexlab{b}}.
\newblock \href {https://doi.org/10.18653/v1/P19-1580} {Analyzing multi-head
  self-attention: Specialized heads do the heavy lifting, the rest can be
  pruned}.
\newblock In \emph{Proceedings of the 57th Annual Meeting of the Association
  for Computational Linguistics}, pages 5797--5808, Florence, Italy.
  Association for Computational Linguistics.

\bibitem[{Wang et~al.(2018)Wang, Zhang, Zhai, Xu, and
  Zong}]{wang-etal-2018-three}
Yining Wang, Jiajun Zhang, Feifei Zhai, Jingfang Xu, and Chengqing Zong. 2018.
\newblock \href {https://doi.org/10.18653/v1/D18-1326} {Three strategies to
  improve one-to-many multilingual translation}.
\newblock In \emph{Proceedings of the 2018 Conference on Empirical Methods in
  Natural Language Processing}, pages 2955--2960, Brussels, Belgium.
  Association for Computational Linguistics.

\bibitem[{Zhang et~al.(2021)Zhang, Bapna, Sennrich, and Firat}]{zhang2021share}
Biao Zhang, Ankur Bapna, Rico Sennrich, and Orhan Firat. 2021.
\newblock \href {https://openreview.net/forum?id=Wj4ODo0uyCF} {Share or not?
  learning to schedule language-specific capacity for multilingual
  translation}.
\newblock In \emph{International Conference on Learning Representations}.

\end{thebibliography}
\bibliographystyle{acl_natbib}

\clearpage
\appendix

\begin{table}[]
    \centering
    \begin{tabular}{l l | l l}
    \toprule
    Code & Name & Code & Name \\
    \midrule
ar & Arabic & ku & Kurdish \\
az & Azerbaijani & lt & Lithuanian \\
be & Belarusian & mk & Macedonian \\
bg & Bulgarian & mn & Mongolian \\
bn & Bengali & mr & Marathi \\
bs & Bosnian & ms & Malay \\
cs & Czech & my & Burmese \\
da & Danish & nb & Norwegian Bokmål \\
de & German & nl & Dutch \\
el & Greek & pl & Polish \\
eo & Esperanto & pt & Portuguese \\
es & Spanish & pt-br & Portuguese \\
et & Estonian & ro & Romanian \\
eu & Basque & ru & Russian \\
fa & Persian & sk & Slovak \\
fi & Finnish & sl & Slovenian \\
fr & French & sq & Albanian \\
fr-ca & French & sr & Serbian \\
gl & Galician & sv & Swedish \\
he & Hebrew & ta & Tamil \\
hi & Hindi & th & Thai \\
hr & Croatian & tr & Turkish \\
hu & Hungarian & uk & Ukrainian \\
hy & Armenian & ur & Urdu \\
id & Indonesian & vi & Vietnamese \\
it & Italian & zh & Chinese \\
ja & Japanese & zh-cn & Chinese \\
ka & Georgian & zh-tw & Chinese \\
    \bottomrule
    \end{tabular}
    \caption{Languages in the Ted Talk Dataset}
    \label{tab:languages}
\end{table}

\section{Correlation of Head Scores} \label{app:correlation}
Here we detail the computation of the correlation of head scores for two pairs of languages $(l_a, l_b)$ and $(l_c, l_d)$. The steps are as follow:
\begin{enumerate}
    \item The two language pairs' head importance scores  $H(l_a, l_b)$ and $H(l_c, l_d)$ are estimated with Equation~\ref{eq:head-score}. Since there are many heads in a Transformer model, both $H(l_a, l_b)$ and $H(l_c, l_d)$ are vectors.
    \item We flatten the scores in $H(l_a, l_b)$ and $H(l_c, l_d)$ into two arrays of scalars. We treat the two arrays as the observations of two variables. Then, we use Spearman correlation to compute the correlation between the two variables. In other words, the input of the Spearman correlation function is the two arrays.
\end{enumerate}

\begin{table*}[]
    \centering
    \begin{tabular}{c  c  c  c  c  c}
        \toprule
        Model   & Lang. Pair &  $H^{enc}$ & $H^{dec}$ & $H^{cross}$ & $H^{self}$\\
        \midrule
All-En  &  X-En & .871 (.086) & .973 (.023) & .978 (.024) & .959 (.024) \\
En-All  &  En-X & .806 (.153) & .720 (.150) & .662 (.204) & .771 (.115) \\
All-All &  X-En & .898 (.073) & .967 (.029) & .980 (.018) & .948 (.046) \\
All-All &  En-X & .813 (.126) & .762 (.141) & .677 (.236) & .810 (.101) \\
        \bottomrule
    \end{tabular}
    \caption{Correlation between the attention head scores when estimated using different language pairs. $H^{cross}$ is the scores for heads across the encoder and the decoder, and $H^{self}$ is the scores for the self-attention head in the decoder.}
    \label{tab:corr2}
\end{table*}

\begin{table*}[]
    \centering
    \begin{tabular}{c  c  c  c  c  c}
        \toprule
        Model   & Lang. Pair &  $H^{enc}$ & $H^{dec}$ & $H^{cross}$ & $H^{self}$\\
        \midrule
All-En  & X-En & .683 (.190) & .925 (.064) & .886 (.099) & .959 (.024) \\
En-All  & En-X & .839 (.187) & .679 (.145) & .585 (.207) & .771 (.115) \\
All-All & X-En & .704 (.169) & .803 (.124) & .787 (.129) & .948 (.046) \\
All-All & En-X & .664 (.213) & .690 (.160) & .545 (.216) & .810 (.101) \\
        \bottomrule
    \end{tabular}
    \caption{The results of comparing language pairs by comparing their top-10 most important attention heads. Let $S_{(a,b)}$ and $S_{(c, d)}$ be the top-10 most important heads for language pair $(l_a, l_b)$, and $S_{(c, d)}$ respectively. We calculate the F1 score between $S_{(a,b)}$ and $S_{(c, d)}$ to measure their similarity. The number in the table is the average F1 scores.}
    \label{tab:corr3}
\end{table*}

\begin{figure*}
    \centering
    \includegraphics[width=0.7\linewidth]{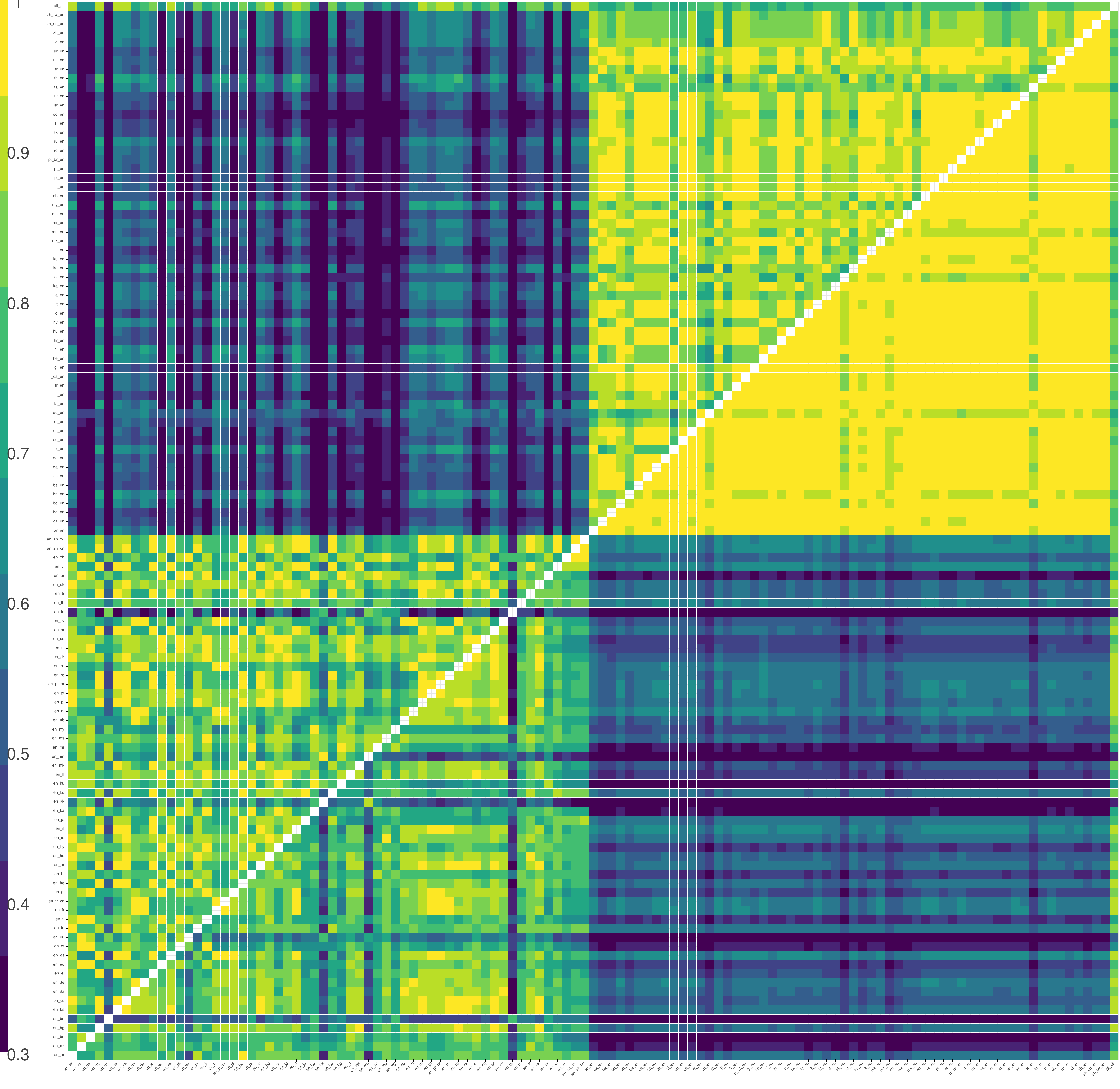}
    \includegraphics[width=0.49\linewidth]{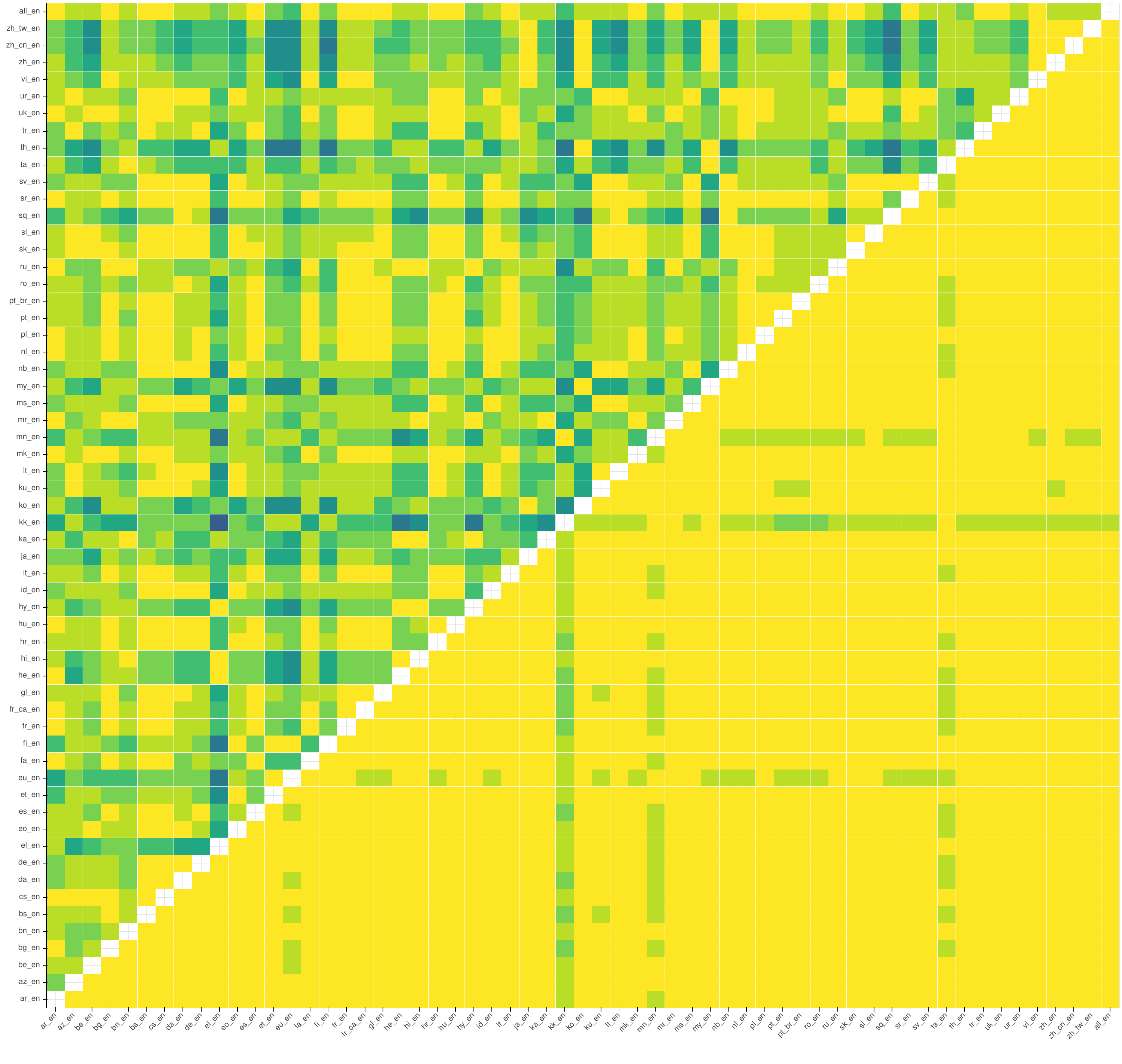}
    \includegraphics[width=0.49\linewidth]{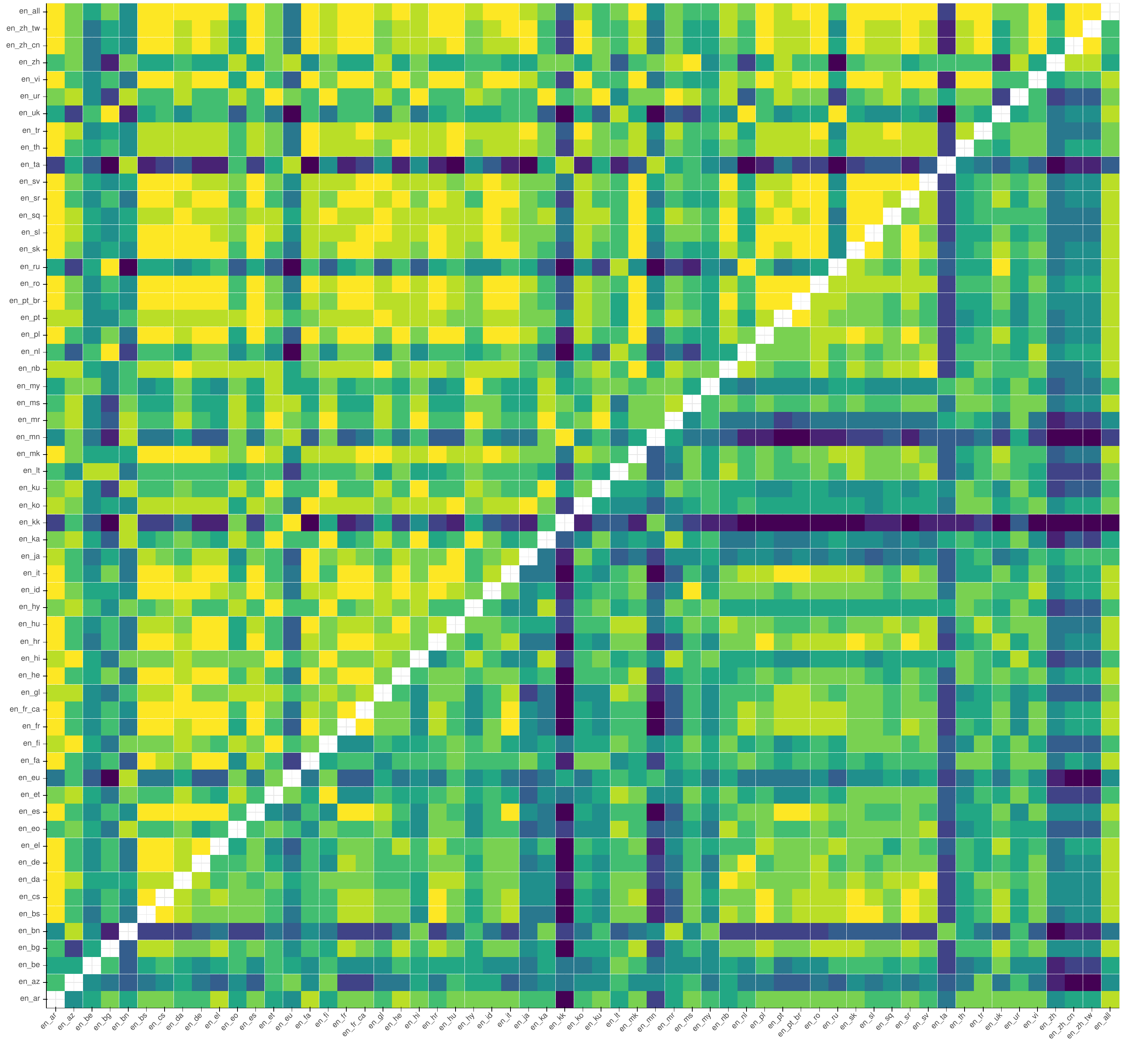}
    \caption{Correlation matrix between language pairs. The top-left corner is the correlation between the encoder head scores $H^{enc}$, while the bottom-right corner is the correlation between the decoder head scores $H^{dec}$. The top matrix is the correlation matrix of the All-All model, while the bottom-left and the bottom-right ones are the correlation matrices of the All-En and the En-All models respectively.}
    \label{fig:my_label}
\end{figure*}


\begin{figure}
    \centering
    \includegraphics[width=0.99\linewidth]{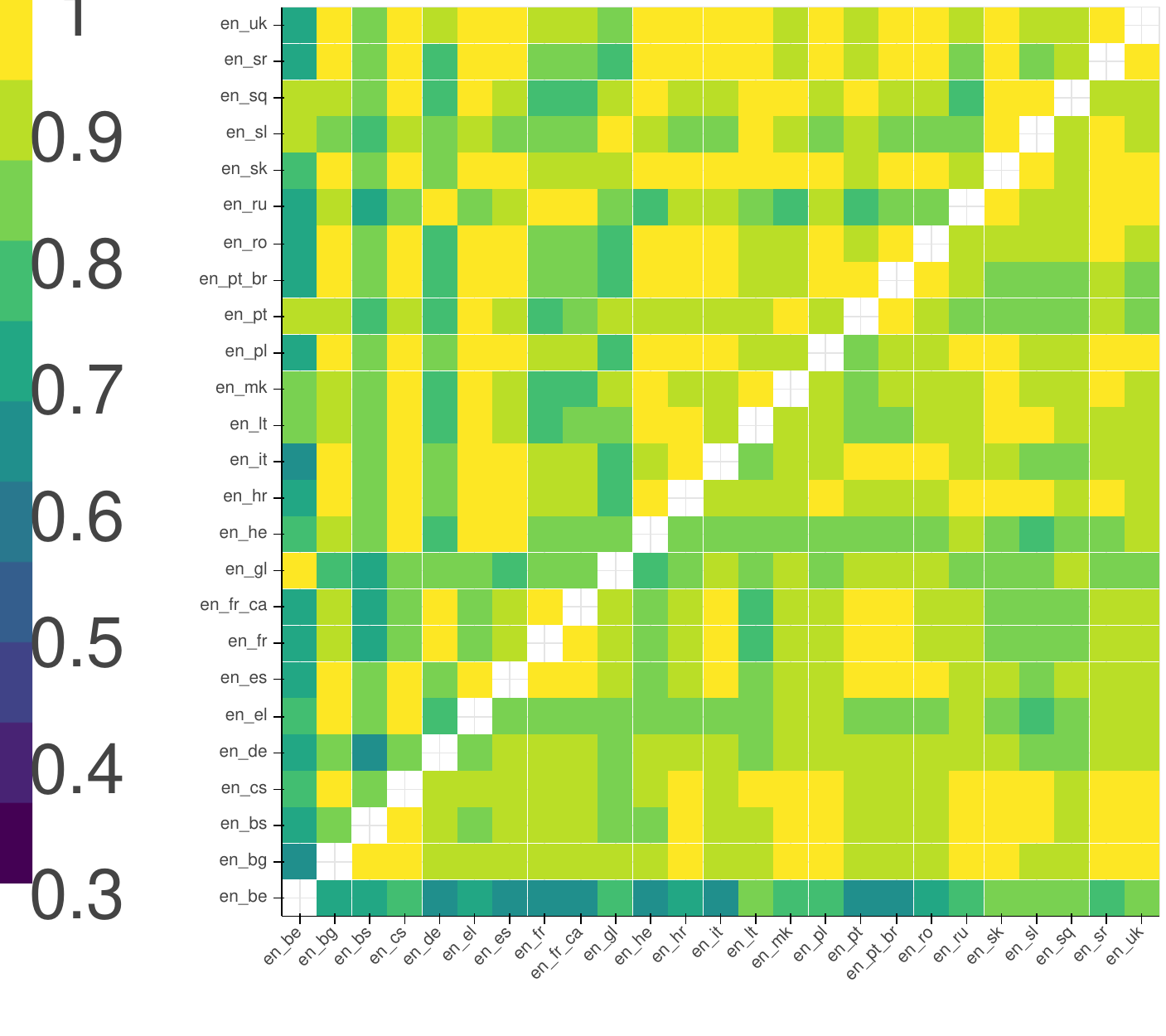}
    \includegraphics[width=0.99\linewidth]{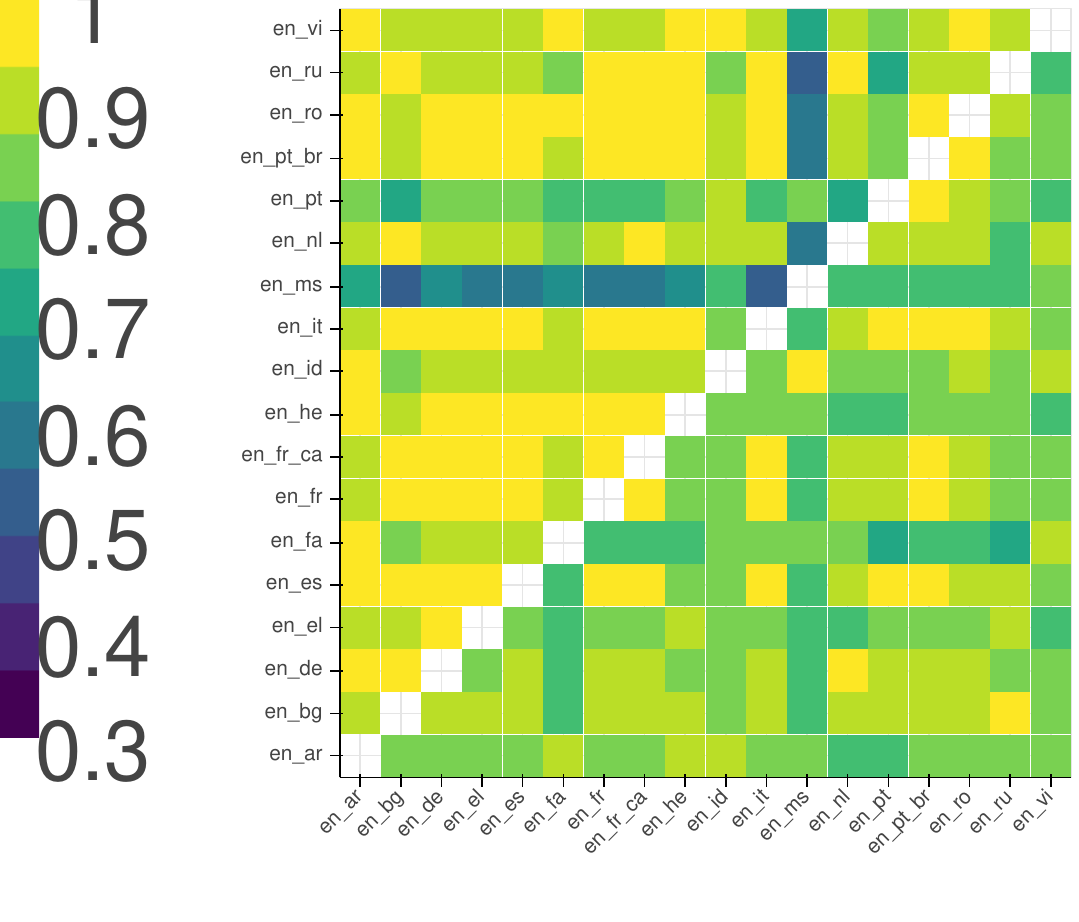}
    \includegraphics[width=0.99\linewidth]{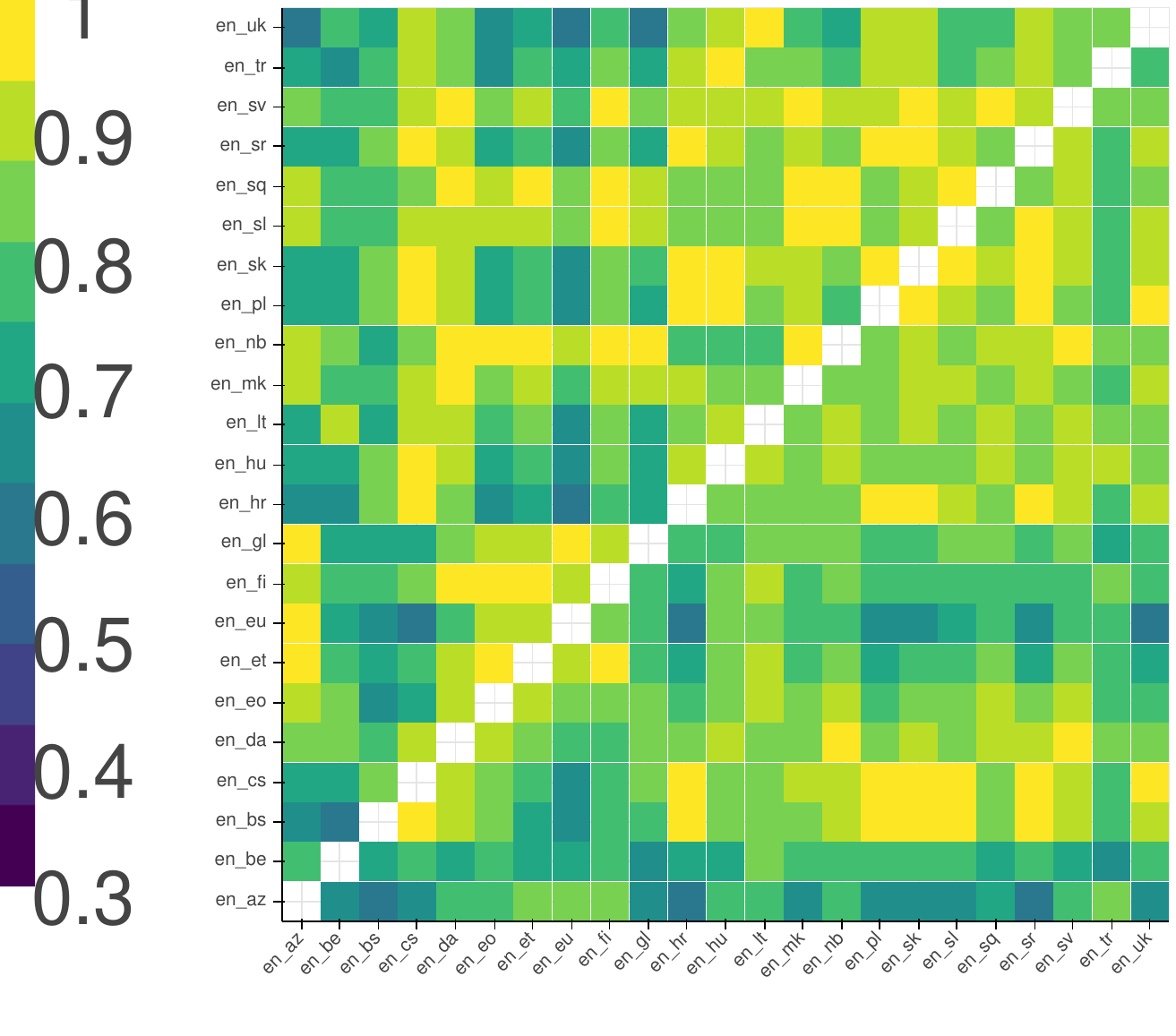}
    \caption{Correlation matrix between language pairs after fine-tuning on the languages clusters. The first figure is the matrix of the fine-tuned All-All model.
    The second and the third ones are the matrix of the En-All model fine-tuned on the language clusters containing the high-resource and the LRL respectively.
    The top-left corner is the correlation between the encoder head scores $H^{enc}$, while the bottom-right corner is the correlation between the decoder head scores $H^{dec}$.}
    \label{fig:my_label}
\end{figure}

\section{Related Related Language Pairs}
The related language pairs used in Section~\ref{sec:improve} are: en-zh\_cn en-it en-es en-vi en-zh\_tw en-nl en-fr en-fr\_ca en-th en-pt\_br en-ru.

\section{Language Clusters}

En-All model:
\begin{itemize}
    \item en-ja en-ko en-zh en-zh-cn en-zh-tw
    \item en-az en-be en-bs en-cs en-da en-eo en-et en-eu en-fi en-gl en-hr en-hu en-lt en-mk en-nb en-pl en-sk en-sl en-sq en-sr en-sv en-tr en-uk
    \item en-bn en-hi en-hy en-ka en-ku en-mr en-my en-ta en-th en-ur
    \item en-ar en-bg en-de en-el en-es en-fa en-fr en-fr-ca en-he en-id en-it en-ms en-nl en-pt en-pt-br en-ro en-ru en-vi
    \item en-kk en-mn
\end{itemize}
All-All:
\begin{itemize}
    \item en-be, en-bg, en-bs, en-cs, en-de, en-el, en-es, en-fr, en-fr-ca, en-gl, en-he, en-hr, en-it, en-lt, en-mk, en-pl, en-pt, en-pt-br, en-ro, en-ru, en-sk, en-sl, en-sq, en-sr, en-uk
    \item en-ar, en-fa, en-ja, en-ko, en-th, en-vi, en-zh, en-zh-cn, en-zh-tw
    \item en-bn, en-hi, en-hy, en-ka, en-ku, en-mr, en-my, en-ur
    \item en-az, en-da, en-eo, en-et, en-fi, en-hu, en-id, en-ms, en-nb, en-nl, en-sv, en-tr
    \item en-eu, en-kk, en-mn, en-ta
\end{itemize}

\section{Random Clusters}
\begin{itemize}
    \item en-pt en-fa en-fr en-kk en-hi en-da en-hu en-de en-nl en-ar en-hy en-zh-cn
    \item en-sr en-fi en-be en-ko en-ru en-ur en-it en-id en-el en-eu en-sq en-zh en-bs en-bn en-sv en-bg en-my en-ro en-ta en-sl en-et en-ku en-mn en-uk en-he en-tr
    \item en-mk en-mr
    \item en-ms en-pl en-pt-br en-cs en-zh-tw en-es
    \item en-vi en-eo en-hr en-nb en-fr-ca en-az en-sk en-ka en-lt en-th en-ja en-gl 
\end{itemize}
Theses random clusters are generated by (1) shuffling the 59 languages, (2) randomly selecting positions. The results 5 segments separated by the 4 positions are the 5 clusters.

\section{Closest Languages}
The closest languages used in Section~\ref{sec:lang-cluster} are:
\begin{itemize}
    \item Az: en-az en-eu en-fi en-tr
    \item Be: en-be en-it en-uk
    \item Gl: en-gl en-pt en-es en-lt en-it en-pt\_br
\end{itemize}

\section{Experimental Details}
\begin{itemize}
    \item Infrastructure: All the experiments can be conducted on one single RTX 2080Ti GPU.
    \item Evaluation: We report the BLEU score calculated by FairSeq.
    \item Version of FairSeq: We use v0.10.0 (\url{https://github.com/pytorch/fairseq/tree/v0.10.0})
    \item Dataset: It can be downloaded from \url{https://github.com/neulab/word-embeddings-for-nmt}.
\end{itemize}

\end{document}